\documentclass[10pt,twocolumn,letterpaper]{article}

\usepackage{cvpr}
\usepackage{times}
\usepackage{epsfig}
\usepackage{graphicx}
\usepackage{amsmath}
\usepackage{amssymb}

\usepackage{hyperref}
\usepackage{amsfonts} 
\usepackage{booktabs}
\usepackage{multirow}
\usepackage{threeparttable}
\usepackage[table,xcdraw]{xcolor}

\begin{document}

\title{Biometric Authentication Based on Enhanced Remote Photoplethysmography Signal Morphology}

\author{Zhaodong Sun\textsuperscript{1}, Xiaobai Li\textsuperscript{2,1}\thanks{Corresponding author.}, Jukka Komulainen\textsuperscript{1}, and Guoying Zhao\textsuperscript{1} \\
\textsuperscript{1}{Center for Machine Vision and Signal Analysis, University of Oulu, Oulu, Finland}\\
\textsuperscript{2}{State Key Laboratory of Blockchain and Data Security, Zhejiang University, Hangzhou, China}\\ 
{\tt\small \{zhaodong.sun, jukka.komulainen, guoying.zhao\}@oulu.fi}, {\tt\small xiaobai.li@zju.edu.cn}\\
}

\maketitle
\thispagestyle{empty}

\begin{abstract}
Remote photoplethysmography (rPPG) is a non-contact method for measuring cardiac signals from facial videos, offering a convenient alternative to contact photoplethysmography (cPPG) obtained from contact sensors. Recent studies have shown that each individual possesses a unique cPPG signal morphology that can be utilized as a biometric identifier, which has inspired us to utilize the morphology of rPPG signals extracted from facial videos for person authentication. Since the facial appearance and rPPG are mixed in the facial videos, we first de-identify facial videos to remove facial appearance while preserving the rPPG information, which protects facial privacy and guarantees that only rPPG is used for authentication. The de-identified videos are fed into an rPPG model to get the rPPG signal morphology for authentication. In the first training stage, unsupervised rPPG training is performed to get coarse rPPG signals. In the second training stage, an rPPG-cPPG hybrid training is performed by incorporating \emph{external} cPPG datasets to achieve rPPG biometric authentication and enhance rPPG signal morphology. Our approach needs only de-identified facial videos with subject IDs to train rPPG authentication models. The experimental results demonstrate that rPPG signal morphology hidden in facial videos can be used for biometric authentication. The code is available at \url{https://github.com/zhaodongsun/rppg_biometrics}.
\end{abstract}


\section{Introduction}


Facial videos contain invisible skin color changes induced by remote photoplethysmography (rPPG) signals, providing valuable cardiovascular information, such as heart rate. Similar to rPPG, contact photoplethysmography (cPPG) captures color changes in fingertips to monitor blood volume changes. cPPG signals, obtained using contact sensors, have been used for biometric authentication \cite{hwang2020evaluation,hwang2021variation}. Given the similar nature and measurement principles of rPPG and cPPG \cite{mcduff2014morph}, rPPG has the potential for biometric authentication. However, the feasibility of rPPG biometric authentication still needs to be validated. Hence, our research questions are: 1) Can rPPG signals be employed for biometric authentication? 2) If so, how can an rPPG-based biometric system be developed? 3) What are the advantages associated with utilizing rPPG biometrics?

\begin{figure}[t]
\centering
\begin{minipage}[b]{0.9\linewidth}
  \centering
  \centerline{\includegraphics[width=\linewidth]{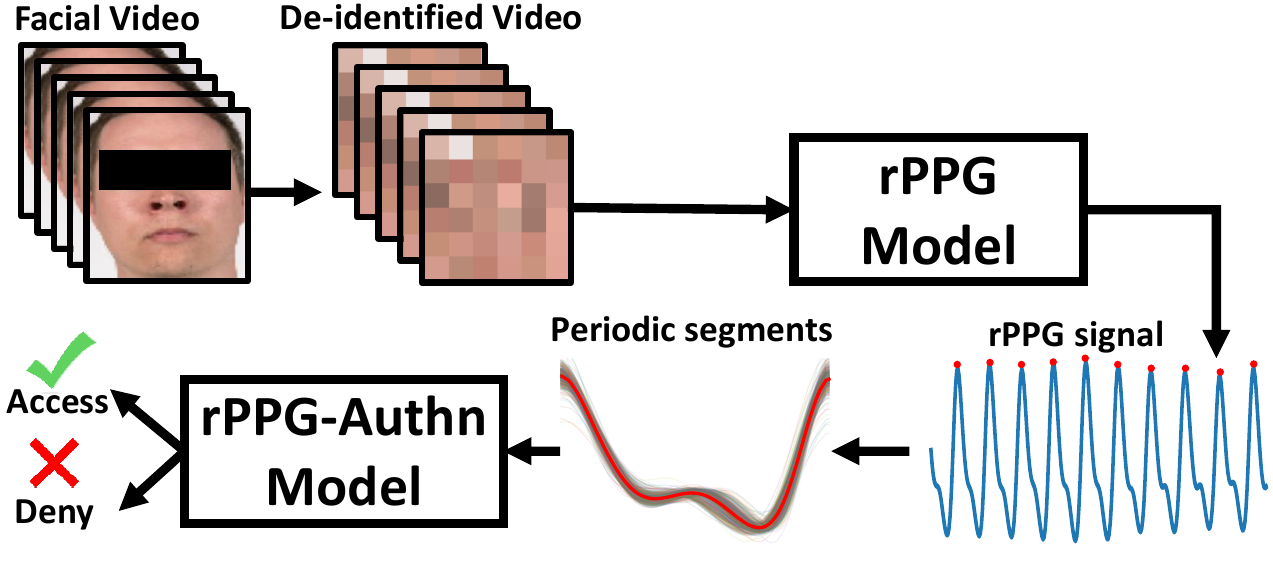}}
  \centerline{(a) rPPG Authentication System}
\end{minipage}

\begin{minipage}[b]{0.9\linewidth}
  \centering
  \centerline{\includegraphics[width=\linewidth]{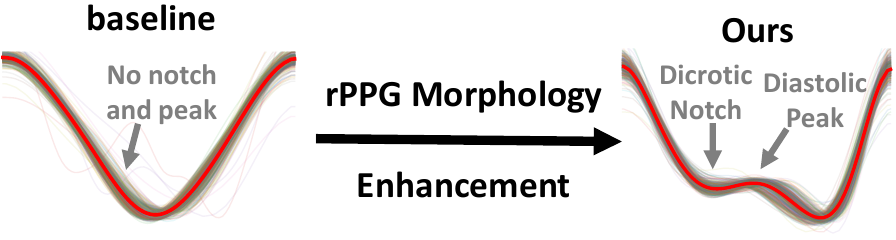}}
  \centerline{(b) rPPG Morphology Enhancement}
\end{minipage}

\caption{(a) rPPG Authentication System. (b) Our method can improve rPPG morphology information. The fiducial points \protect\cite{lovisotto2020seeing} like the systolic peaks and diastolic peaks are the main subject-specific biometric characteristics in rPPG signals.}
\label{fig:intro}
\vspace{-0.3cm}
\end{figure}

We first examine the quality and discriminative power of rPPG signals. rPPG signals are derived from subtle changes in facial color caused by blood volume changes during heartbeats. Recent advances \cite{yu2019remoteBMVC,NEURIPS2020_e1228be4} have achieved high-quality rPPG measurement, especially when the face has minimal or no movement. Hence, it is feasible to obtain high-quality rPPG signals. However, the question remains whether these high-quality rPPG signals contain subject-specific biometric characteristics. One work \cite{patil2018non} has tried using rPPG for biometrics, but the preliminary study was limited by a small-scale dataset and low-quality rPPG, offering inadequate authentication performance for practical applications.

In this paper, we propose an rPPG-based method for biometric authentication, as shown in Fig. \ref{fig:intro}(a). Considering facial appearance and rPPG are mixed together in facial videos, we first de-identify facial videos while preserving the rPPG information. This step can guarantee that only rPPG information is used for biometric authentication while facial appearance cannot be used. In addition, this step can also conceal sensitive facial appearance information for privacy protection. The first module is the rPPG model that can extract rPPG signals from the de-identified facial videos. The second module is the rPPG-Authn model that utilizes the rPPG morphology to output person authentication results. We design a two-stage training strategy and rPPG-cPPG hybrid training by incorporating \emph{external} cPPG datasets to exploit rPPG morphology for biometric authentication. Fig. \ref{fig:intro}(b) illustrates the rPPG morphology enhancement. Note that we only use de-identified videos with subject IDs for rPPG biometrics.

There are several advantages of rPPG biometrics. Compared with facial appearances, the rPPG biometric system only utilizes de-identified facial videos, eliminating the need for sensitive facial appearance. Moreover, rPPG biometrics offers an additional degree of resistance to spoofing, as rPPG inherently serves as a countermeasure to presentation attacks \cite{liu2018learning,liu2022learning}. In contrast, without dedicated presentation attack detection (PAD) methods, conventional face recognition algorithms are vulnerable to presentation attacks and less secure than rPPG-based biometrics. Additionally, since both rPPG biometrics and face recognition use facial videos as data sources, combining both biometric modalities can potentially enhance both accuracy and security. When compared with cPPG biometrics, rPPG biometrics offers the advantages of being non-contact and only requiring off-the-shelf cameras, while cPPG biometrics necessitates specific contact sensors like pulse oximeters. Compared with iris recognition \cite{wildes1997iris,daugman2009iris} which requires iris scanners, rPPG biometrics only requires cheap RGB cameras and is robust to presentation attacks.


Our contributions include: 
\begin{enumerate}
    \item We propose a new biometric authentication method based on rPPG. We utilize two-stage training to achieve rPPG morphology enhancement and accurate biometric authentication performance. We illustrate that utilizing de-identified facial videos is effective for rPPG biometric authentication and ensures the protection of facial appearance privacy.
    \item We conduct comprehensive experiments on multiple datasets to validate the discriminative power of rPPG biometrics. We demonstrate that rPPG biometrics can achieve comparable performance with cPPG biometrics. We also investigate factors that may influence the performance of rPPG biometrics.
    \item We discover that our rPPG-based biometric method can enhance rPPG morphology, which opens up possibilities for rPPG morphology learning from facial videos.
\end{enumerate}

\section{Related Work}

\subsection{rPPG Measurement}

\cite{verkruysse2008remote} initially proposed measuring rPPG from face videos via the green channel. Subsequent handcrafted methods have been introduced to enhance the quality of the rPPG signal \cite{poh2010advancements,de2013robust,li2014remote,tulyakov2016self,wang2014exploiting}. Recently, there has been rapid growth in deep learning (DL) approaches for rPPG measurement. Several studies \cite{chen2018deepphys,vspetlik2018visual,NEURIPS2020_e1228be4,nowara2021benefit,li2023learning} utilize 2D convolutional neural networks (CNN) to input consecutive video frames for rPPG measurement. Another set of DL-based methods \cite{niu2019rhythmnet,niu2020video,lu2021dual,lu2023neuron,du2023dual} employ a spatial-temporal signal map obtained from different facial regions, which is then fed into 2DCNN models. 3DCNN-based methods \cite{yu2019remote} and transformer-based methods \cite{yu2022physformer,yu2023physformer++} have been proposed to enhance spatiotemporal performance and long-range spatiotemporal perception. 

Additionally, multiple unsupervised rPPG methods \cite{gideon2021way,wang2022self,sun2022contrast,speth2023non,yang2022simper,yue2023facial} have been proposed. Since GT signals are expensive to collect and synchronize in rPPG datasets, unsupervised rPPG methods only require facial videos for training without any GT signal and achieve performance similar to the supervised methods. However, most works on rPPG measurement primarily focus on the accuracy of heart rate estimation, while neglecting the rPPG morphology.

\subsection{cPPG-based Biometrics}
\cite{gu2003novel} was the first attempt to utilize cPPG for biometric authentication. They extracted some fundamental morphological features, such as peak upward/downward slopes, for cPPG biometrics. Subsequently, other studies have explored additional morphological features, including cPPG derivatives \cite{yao2007pilot} and fiducial points \cite{lovisotto2020seeing}. More recently, researchers have focused on employing DL methods to automatically extract morphological features. \cite{luque2018end,biswas2019cornet,lee2020cross} directly input cPPG signals into 1DCNN or long short-term memory (LSTM) architectures to conduct biometric authentication, while \cite{hwang2020evaluation,hwang2021variation} cut cPPG signals into periodic segments and utilize multiple representations of these periodic segments as inputs to a 1DCNN model. Furthermore, \cite{hwang2020evaluation} has collected datasets for cPPG biometrics and investigated the permanence of cPPG biometrics. There exists one preliminary work on rPPG biometrics \cite{patil2018non}, but only a traditional independent component analysis (ICA) based method \cite{poh2010advancements} was applied for rPPG extraction, which yields low-quality rPPG morphology for biometric authentication.

\section{Method}

Our method consists of facial video de-identification and two training stages. As the rPPG signal does not rely on facial appearance, we first de-identify the input video to avoid facial appearance being used by our method. In the first training stage, we perform unsupervised rPPG training on the de-identified videos to achieve basic rPPG signal measurement. In the second training stage, we use rPPG-cPPG hybrid training for biometric authentication and rPPG morphology enhancement.

\subsection{Face De-identification for rPPG Biometrics}

We propose to de-identify facial videos using spatial downsampling and pixel permutation. This step aims to obfuscate facial appearances while preserving the rPPG information. Since rPPG signals are spatially redundant at different facial regions and largely independent of spatial information as shown by \cite{tulyakov2016self,niu2018synrhythm}, rPPG signals can be well preserved in this step while facial appearances are completely erased. The reasons for face de-identification are twofold. First, the facial appearance and rPPG information are intertwined in facial videos. We remove facial appearance to make sure that the biometric model performs recognition solely based on the rPPG information. Second, this step can remove facial appearances to protect facial privacy information during rPPG authentication.

\begin{figure}[t]
\centering
\begin{minipage}[b]{0.9\linewidth}
  \centering
  \centerline{\includegraphics[width=\linewidth]{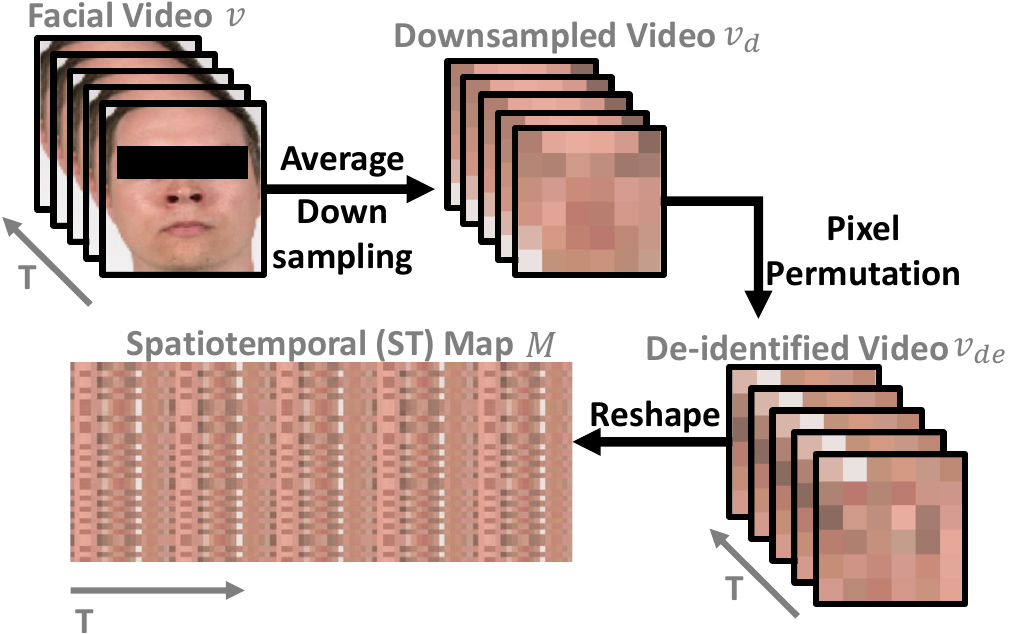}}
\end{minipage}
\caption{Face de-identification for rPPG biometrics. The facial appearance is obfuscated while rPPG information is retained.}
\label{fig:de-id}
\vspace{-0.3cm}
\end{figure}

The facial video is de-identified as shown in Fig. \ref{fig:de-id}. Faces in the original videos are cropped using OpenFace \cite{baltrusaitis2018openface} by locating the boundary landmarks. The cropped facial video $v \in \mathbb{R}^{T \times H \times W \times 3}$, where $T$, $H$, and $W$ are time length, height, and width, is downsampled by averaging the pixels in a sample region to get $v_{d} \in \mathbb{R}^{T \times 6 \times 6 \times 3}$. It has been demonstrated that such downsampled facial videos are still effective in rPPG estimation \cite{tulyakov2016self,niu2018synrhythm}. Since rPPG signal extraction does not largely depend on spatial information \cite{tulyakov2016self}, we further permutate the pixels to completely obfuscate the spatial information to get $v_{de} \in \mathbb{R}^{T \times 6 \times 6 \times 3}$. Note that the permutation pattern is the same for each frame in a video but distinct for different videos. Since the spatial information is eliminated, we reshape the de-identified video $v_{de}$ into a spatiotemporal (ST) map $M \in \mathbb{R}^{36 \times T \times 3}$ for compact rPPG representation like \cite{niu2018synrhythm}.

\subsection{The 1st training stage: rPPG Unsupervised Pre-training}
This stage aims to train a basic rPPG model capable of extracting rPPG with precise heartbeats. We use unsupervised training to obtain the basic rPPG model. The main reasons for unsupervised training are: 1) Unsupervised rPPG training does not require GT PPG signals from contact sensors, which means only facial videos with subject IDs are required in our entire method. 2) The performance of unsupervised rPPG training \cite{gideon2021way,sun2022contrast} is on par with supervised methods.


\begin{figure}[t]
\centering
\begin{minipage}[b]{0.8\linewidth}
  \centering
  \centerline{\includegraphics[width=\linewidth]{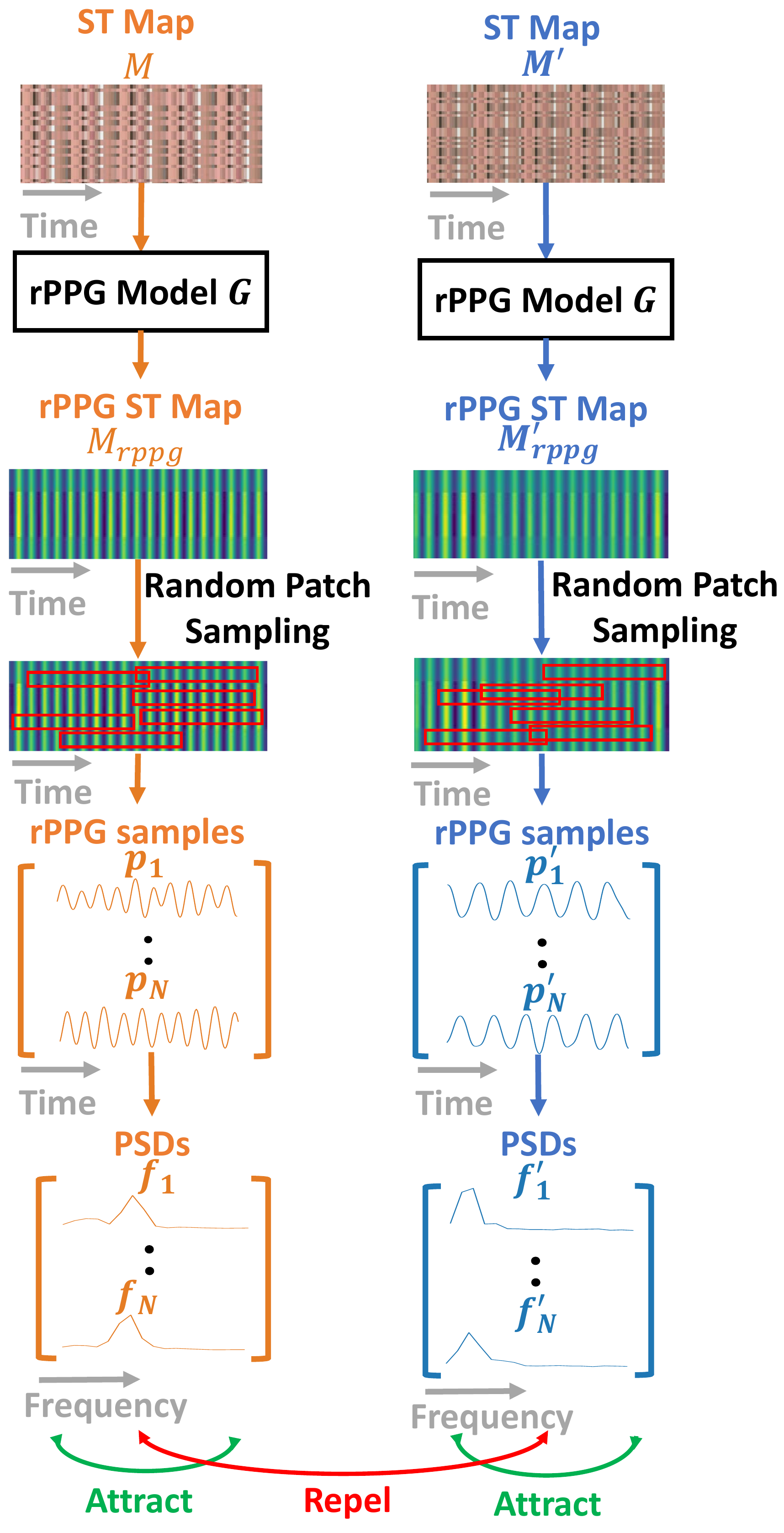}}
\end{minipage}
\caption{The diagram of Contrast-Phys-2D (CP2D) for rPPG unsupervised pre-training based on contrastive learning.}
\label{fig:unsupervised}
\end{figure}

We adopt and customize the unsupervised Contrast-Phys (CP) architecture \cite{sun2022contrast} to 2D ST-map inputs since CP can only use face videos as inputs. The modified method called Contrast-Phys-2D (CP2D) is shown in Fig. \ref{fig:unsupervised}. Two different ST maps $M, M' \in \mathbb{R}^{36 \times T \times 3}$ from two different videos are the inputs of the rPPG model $G$, where $T$ is 10 seconds. The rPPG model $G$ is based on a 2D convolutional neural network to output rPPG ST maps $M_{rppg}, M'_{rppg} \in \mathbb{R}^{S \times T}$ where rPPG signals are stacked vertically. Similar to CP, the spatial dimension $S$ is set as four. The architecture of the rPPG model $G$ is presented in the supplementary materials. Inspired by spatiotemporal rPPG sampling in CP, we use a patch with the shape $(1, T/2)$ to randomly get $N=16$ rPPG ST samples $\{m_{rppg(1)},...,m_{rppg(N)}\}, \{{m'_{rppg(1)},...,m'_{rppg(N)}}\}$ from rPPG ST maps $M_{rppg}, M'_{rppg}$, respectively. The rPPG ST samples are averaged along the spatial dimension to get rPPG samples $\{p_1,...,p_N\}, \{p'_1,...,p'_N\}$ and the corresponding power spectral densities (PSDs) $\{f_1,...,f_N\}, \{f'_1,...,f'_N\}$. We use rPPG prior knowledge \cite{sun2022contrast} including rPPG spatiotemporal similarity and cross-video rPPG dissimilarity to make positive pairs ($(f_i, f_j)$ or $(f'_i, f'_j)$) and negative pairs $(f_i, f'_j)$, which can be used in the positive and negative terms in the contrastive loss $L$. The contrastive loss $L$ is used to pull together the PSDs originating from the same videos and push away the PSDs from different videos. The loss function $L$ is shown below. During inference, the rPPG ST map $M_{rppg} \in \mathbb{R}^{S \times T}$ is averaged along the spatial dimension to get the rPPG signal $s_{rppg} \in \mathbb{R}^T$.

\begin{equation}
\resizebox{0.47\textwidth}{!}{$L = \sum\limits_{i=1}^{N} \sum\limits_{\substack{j=1 \\j \neq i}}^{N} \frac{\parallel f_i - f_j \parallel^2 + \parallel f'_i - f'_j \parallel^2}{2N(N-1)} - \sum\limits_{i=1}^{N} \sum\limits_{j=1}^{N} \frac{\parallel f_i - f'_j \parallel^2}{N^2}$}
\end{equation}


\begin{figure}[t]
\centering
\begin{minipage}[b]{0.9\linewidth}
  \centering
  \centerline{\includegraphics[width=\linewidth]{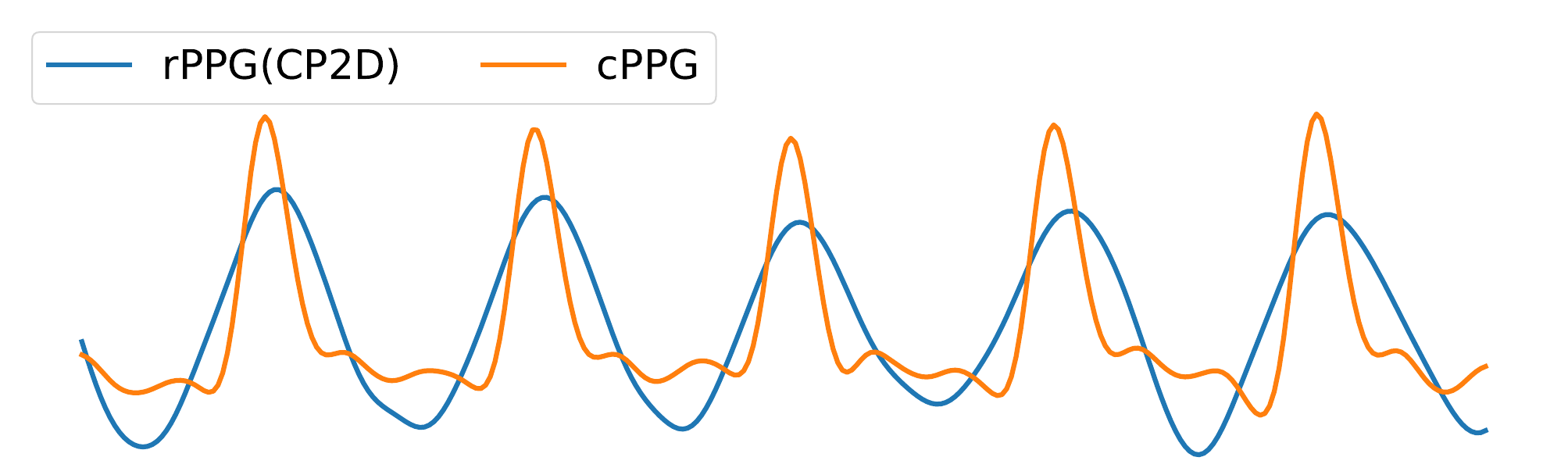}}
\end{minipage}
\caption{GT cPPG signal and rPPG signal extracted by CP2D. After the first training stage, the rPPG signal has accurate heartbeats but lacks morphology information.}
\label{fig:rppg_cp2d_cppg}
\end{figure}

However, since CP2D does not utilize any prior knowledge about morphology, the resulting rPPG signals lack morphology information. Fig. \ref{fig:rppg_cp2d_cppg} shows a GT cPPG signal and an rPPG signal produced by CP2D. CP2D generates an rPPG signal with accurate heartbeats that align with those of the cPPG signal. However, the morphological features, such as the dicrotic notch and diastolic peak evident in the cPPG morphology, are not clearly discernible in the rPPG signals. Since these morphological features play a crucial role in differentiating individuals, we aim to further refine the rPPG signal morphology at the second training stage.

\subsection{The 2nd training stage: rPPG-cPPG Hybrid Training}

\begin{figure*}[t]
\centering
\begin{minipage}[b]{0.95\linewidth}
  \centering
  \centerline{\includegraphics[width=\linewidth]{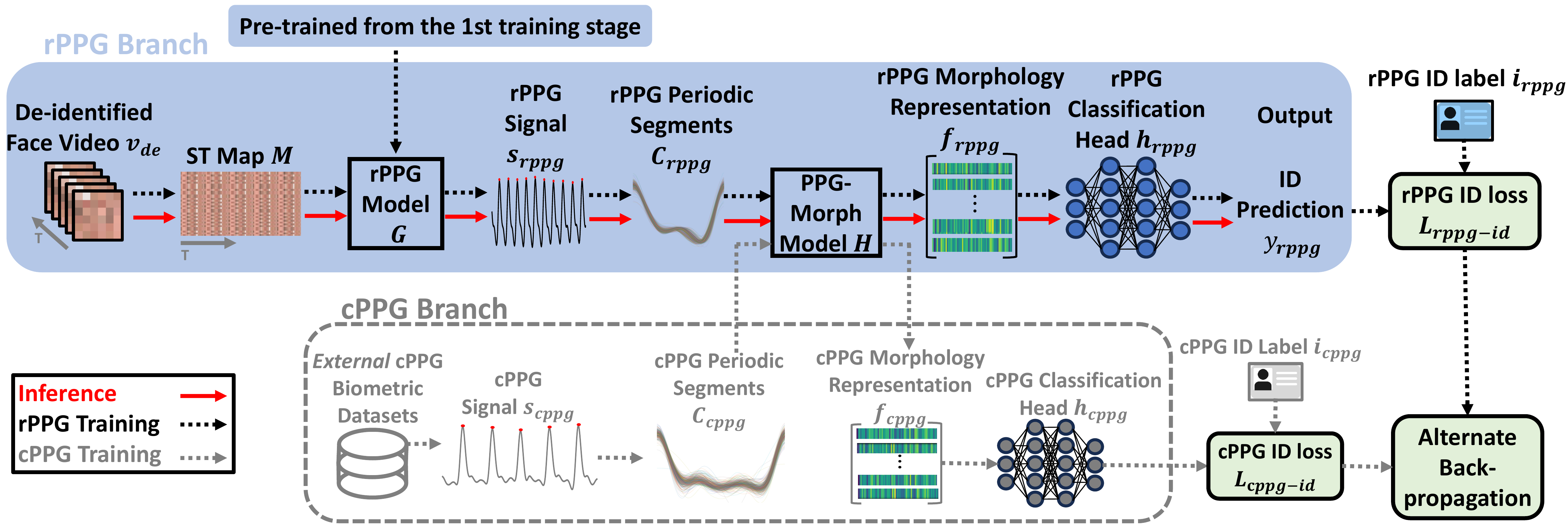}}
\end{minipage}
\caption{rPPG-cPPG hybrid training. The rPPG branch and cPPG branch are trained alternatively to utilize external cPPG signals to enhance the rPPG morphology fully.}
\vspace{-0.3cm}
\label{fig:hybrid}
\end{figure*}

At the second training stage, we further refine rPPG signals to obtain morphology information. Fig. \ref{fig:hybrid} shows the rPPG-cPPG hybrid training, where the rPPG branch utilizes face videos and ID labels during training. On the other hand, the cPPG branch uses \emph{external} cPPG biometric datasets to encourage the PPG-Morph model $H$ to learn morphology information, which can be incorporated into the rPPG branch through the PPG-Morph model $H$. The PPG-Morph model $H$ comprises 1DCNN layers and transformer layers that extract morphological features from periodic segments. The two branches are trained alternately to facilitate the sharing of morphology information between the rPPG and cPPG branches. Note that our method only requires de-identified facial videos with subject IDs during training (enrollment) and only needs de-identified facial videos during inference.

\subsubsection{rPPG Branch}
The rPPG branch can extract rPPG morphology and use it to differentiate individuals. This branch only requires a de-identified facial video $v_{de}$ and the ID label $i_{rppg}$ and does not need any GT cPPG signal for training. Therefore, de-identified facial videos with ID labels are sufficient for enrollment in the proposed rPPG biometrics scheme. The ST map $M$ derived from the de-identified facial video $v_{de}$ is fed into the pre-trained rPPG model $G$ to obtain the rPPG signal $s_{rppg}$. Note that the rPPG model $G$ is the pre-trained model from the first unsupervised training stage. To segment the signal, the systolic peaks are located, and the signal $s_{rppg} \in \mathbb{R}^T$ is divided into K clips. Due to heart rate variability, the K clips may have different lengths, so the clip length is interpolated to 90 in order to obtain rPPG periodic segments. The choice of a length of 90 is based on the fact that the minimum heart rate (40 beats per minute) for a 60 Hz signal produces the longest periodic segment with a length of 90. Consequently, we obtain $C_{rppg} \in \mathbb{R}^{K\times90}$. To predict an authentication score for an individual, we use the PPG-Morph model $H$ and the rPPG classification head $h_{rppg}$, which provides the rPPG morphology representation $f_{rppg} \in \mathbb{R}^{64}$ and ID probability $y_{rppg} \in [0,1]^{K \times N_{rppg}}$, where $N_{rppg}$ is the number of individuals in the rPPG biometric dataset. The cross-entropy loss is used for ID classification, which is

\begin{equation}
L_{rppg-id}(y_{rppg}, i_{rppg})=-\frac{1}{K}\sum_{k=0}^K \log(y_{rppg}^{k,i_{rppg}})
\end{equation}
where $y_{rppg}^{k,i_{rppg}}$ is the predicted probability of the \textit{k}th periodic segment belonging to the ID label $i_{rppg} \in \{1,2,...,N_{rppg}\}$.

\subsubsection{cPPG Branch}
The cPPG branch utilizes \emph{external} cPPG biometric datasets including Biosec2 \cite{hwang2020evaluation}, BIDMC \cite{pimentel2016toward,goldberger2000physiobank}, and PRRB \cite{karlen2013multiparameter}, to learn PPG morphology. Note that the \emph{external} cPPG biometric datasets are available online and are not related to the facial videos in the rPPG branch. Similar to the rPPG branch, the cPPG signal $s_{cppg} \in \mathbb{R}^T$ is processed to obtain $K'$ cPPG periodic segments $C_{cppg} \in \mathbb{R}^{K'\times90}$. The PPG-Morph model $H$ and cPPG classification head $h_{cppg}$ are employed to generate the cPPG morphology representation $f_{cppg} \in \mathbb{R}^{64}$ and the ID probability prediction $y_{cppg} \in [0,1]^{K' \times N_{cppg}}$, where $N_{cppg}$ is the number of individuals in the \emph{external} cPPG biometric datasets. Note that the PPG-Morph model $H$ is shared by both the rPPG branch and cPPG branch, allowing the cPPG branch to transfer the learned morphology information to the rPPG branch. The cross-entropy loss is utilized in this branch, which is

\begin{equation}
L_{cppg-id}(y_{cppg},i_{cppg})=-\frac{1}{K'}\sum_{k=0}^{K'} \log(y_{cppg}^{k,i_{cppg}})
\end{equation}
where $y_{cppg}^{k,i_{cppg}}$ is the predicted probability of the \textit{k}th periodic segment belonging to the ID label $i_{cppg} \in \{1,2,...,N_{cppg}\}$.

\subsubsection{Alternate Backpropagation}
We alternately train the two branches and backpropagate the gradient of the two loss functions $L_{rppg-id}$ and $L_{cppg-id}$ to achieve rPPG-cPPG hybrid training. During the first step, de-identified facial videos and ID labels are sampled from the rPPG biometric dataset to calculate the loss $L_{rppg-id}$, and the gradient of $L_{rppg-id}$ is backpropagated to update the rPPG model $G$, the PPG-Morph model $H$, and the rPPG classification head $h_{rppg}$. During the second step, cPPG signals and ID labels are sampled from \emph{external} cPPG biometric datasets to calculate the loss $L_{cppg-id}$, and the gradient of $L_{cppg-id}$ is backpropagated to update PPG-Morph model $H$ and the cPPG classification head $h_{cppg}$. These two steps are repeated in an alternating manner, allowing the two branches to be trained in turns. The cPPG branch uses external cPPG datasets to encourage the PPG-Morph model $H$ to learn morphology information. The morphology features learned from the cPPG branch can then be incorporated into the rPPG branch since the PPG-Morph model $H$ is shared by both cPPG and rPPG branches thus rPPG features are enhanced. The supplementary materials provide a detailed description of the algorithm.



\section{Experiments}

\begin{table*}[]
\centering
{
\begin{tabular}{@{}lccccc@{}}
\toprule
\multirow{3}{*}{\begin{tabular}[c]{@{}l@{}}\\Signal length\end{tabular}} & \multicolumn{5}{c}{EER$\downarrow$/AUC$\uparrow$} \\ \cmidrule(lr){2-6} 
                               & \multicolumn{2}{c}{OBF}         & UBFC-rPPG       & \multicolumn{2}{c}{PURE}         \\ 
                               \cmidrule(lr){2-3} \cmidrule(lr){4-4} \cmidrule(lr){5-6}
                               & \begin{tabular}[c]{@{}l@{}}intra-session\end{tabular}    & \begin{tabular}[c]{@{}l@{}}cross-session\end{tabular}     & intra-session      & intra-session & cross-session \\ \cmidrule{1-1} \cmidrule(lr){2-3} \cmidrule(lr){4-4} \cmidrule(lr){5-6}
20 heartbeats ($\sim$20 sec) & \bf 0.17\%/99.97\% & \bf 2.16\%/98.10\% & \bf 0\%/100\%    & \bf 0\%/100\%      & \bf 9.59\%/93.70\%  \\ \midrule
10 heartbeats ($\sim$10 sec) & 0.14\%/99.98\% & 2.61\%/98.04\% &  \bf 0\%/100\% & 0.33\%/99.67\% & 14.00\%/91.17\% \\ \midrule
5 heartbeats ($\sim$5 sec) & 0.33\%/99.97\% & 3.81\%/97.89\% &  0.01\%/99.99\% & 0.58\%/99.36\% & 18.32\%/86.81\% \\ \bottomrule
\end{tabular}
\caption{EER and AUC for rPPG authentication on OBF, UBFC-rPPG, and PURE datasets.}
\vspace{-0.3cm}
\label{tab:veri}
}
\end{table*}

\subsection{Implementation Details}

\textbf{Datasets.} We considered three public rPPG datasets, namely OBF \cite{li2018obf}, PURE \cite{stricker2014non}, and UBFC-rPPG \cite{bobbia2019unsupervised}. The scales of these rPPG datasets are enough to validate the feasibility of rPPG biometrics since previous cPPG biometric datasets \cite{hwang2020evaluation,hwang2021variation} also have similar scales. These rPPG datasets consist of facial videos, GT cPPG signals, and ID labels, but our method does not require the GT cPPG. \textbf{OBF} dataset \cite{li2018obf} consists of data from 100 healthy subjects. Two 5-minute RGB facial videos were recorded for each participant. For each subject, the first facial video was recorded at rest, while the second was recorded after exercise. During the recording, participants remained seated without head or facial motions. Videos have a resolution of 1920×1080 at 60 frames per second (fps). \textbf{UBFC-rPPG} dataset \cite{bobbia2019unsupervised} was captured using a webcam at a resolution of 640x480 at 30 fps. In each recording, the subject was positioned 1 meter away from the camera and playing a mathematical game, with the face centrally located within the video frame. The database consists of data from 42 participants, with each one having a 1-minute video. \textbf{PURE} dataset \cite{stricker2014non} contains data from 10 subjects. Face videos for each subject were captured in 6 distinct scenarios: steady, talking, slow translation, fast translation, small rotation, and medium rotation, leading to a total of 60 one-minute RGB videos. Videos have a resolution of 640×480 at 30 fps.




Additionally, we combined the Biosec2 \cite{hwang2020evaluation}, BIDMC \cite{pimentel2016toward,goldberger2000physiobank}, and PRRB \cite{karlen2013multiparameter} datasets to create the \emph{external} cPPG biometric dataset. These datasets contain cPPG signals from 195 subjects for the cPPG branch in the rPPG-cPPG hybrid training. More details about datasets are provided in the supplementary materials.

\textbf{Experimental Setup.} Our rPPG biometric experiments follow the previous cPPG biometric protocol \cite{hwang2020evaluation,hwang2021variation} where the training and test sets have the same persons but might be recorded in the same session (intra-session test) or recorded in different sessions (cross-session test). For the OBF dataset, we divide each pre-exercise video into three parts: the first 60\% length is used for training, the following 20\% length is used for validation, and the last 20\% length is used for intra-session testing. The post-exercise videos are reserved for cross-session testing. As for the UBFC-rPPG dataset, the same division is applied to each video. Since each subject only contributes one video, only intra-session testing can be conducted on this dataset. Moving on to the PURE dataset, the same division is applied to each steady video. The videos involving head motion tasks are used exclusively for cross-session testing. At the first training stage, we select the best rPPG model with the lowest irrelevant power ratio (IPR) in the validation set, as conducted in \cite{gideon2021way,sun2022contrast}. At the second training stage, we choose the best-performing models based on the lowest equal error rate (EER) in the validation set. Both training stages are carried out on a single Nvidia V100 GPU and employ the Adam optimizer with a learning rate of 1e-3. During inference, the predicted probabilities from consecutive periodic segments (5 beats, 10 beats, and 20 beats) are averaged.

\textbf{Evaluation Metrics.} Since the model does multi-class classification, we use the one-vs-rest strategy to get the authentication results for each person. Therefore, each person has a binary classification. For each person, we can change the threshold of the model prediction output for that person to get the binary predictions, and we can plot false positive rates and true positive rates in a graph, which is the receiver operating characteristic (ROC) curve. Areas under curve (AUC) is the area under the ROC curve. If we change the threshold, we can find the threshold where the false positive rate and the false negative rate are equal. The EER is the false positive rate or false negative rate at this threshold. The final EER and AUC are averaged across all subjects. To evaluate the rPPG morphology, we calculate the Pearson correlation between the means of periodic segments from rPPG and the GT cPPG. More details are in the supplementary materials.

\subsection{Results and discussions}
\subsubsection{Results and discussions about rPPG authentication.}
Table \ref{tab:veri} presents the results of rPPG authentication with varying signal lengths. The performance of rPPG authentication improves with longer signal lengths, such as 20 heartbeats, compared to shorter signal lengths like 10 or 5 beats. On all three datasets, the intra-session performance is satisfactory, with EERs below 1\% and AUCs above 99\%. However, the performance decreases during cross-session testing. On the OBF dataset, the cross-session (pre-exercise $\to$ post-exercise) performance is slightly lower than the intra-session (pre-exercise $\to$ pre-exercise) performance, but still achieves EER of 2.16\%. On the PURE dataset, there is a significant drop in performance during cross-session (steady $\to$ motion tasks) compared to intra-session (steady $\to$ steady) due to the adverse impact of motion tasks on the quality of rPPG signals. Conversely, although the OBF dataset includes exercises to increase heart rates, it does not involve facial movements. This indicates that rPPG biometrics is sensitive to low-quality rPPG caused by facial motions but rPPG has reliable and unique biometric information evidenced by the varying heart rates from the same people. In practical usage, users will face the camera and keep still (like face recognition), thus such large intended head motions will not be a concern.

\begin{figure}[t]
\centering
\begin{minipage}[b]{0.95\linewidth}
  \centering
  \centerline{\includegraphics[width=\linewidth]{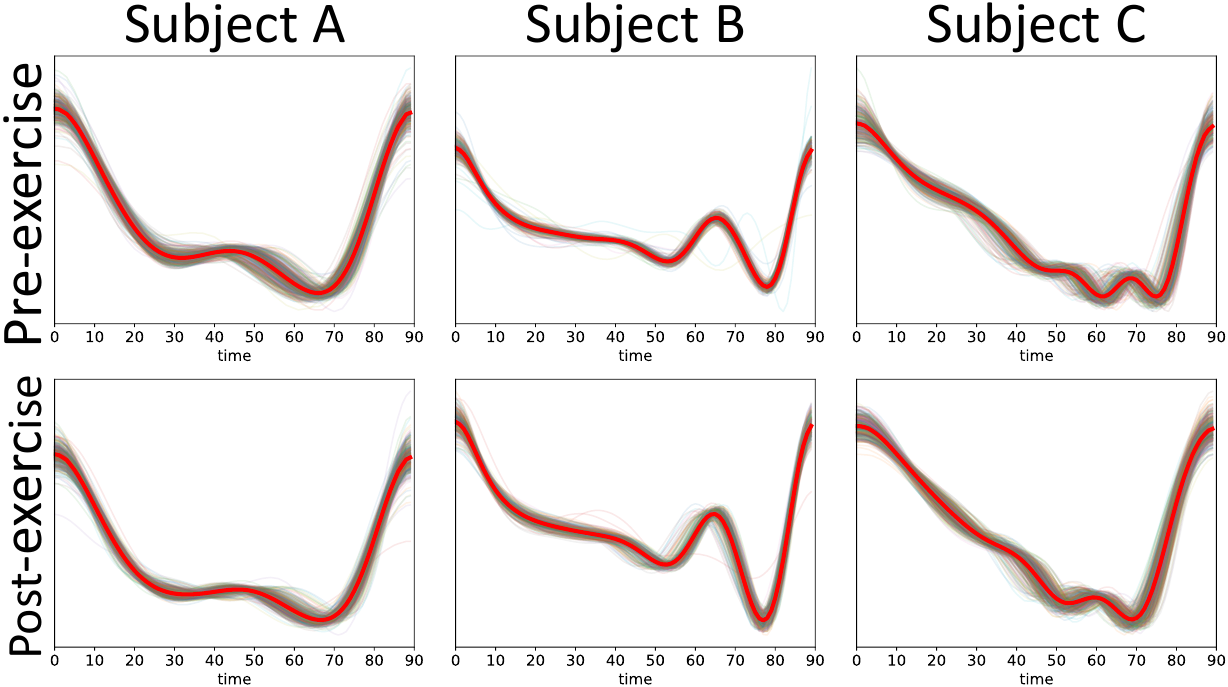}}
  \centerline{(a) rPPG periodic segments from OBF dataset}
\end{minipage}

\begin{minipage}[b]{0.95\linewidth}
  \centering
  \centerline{\includegraphics[width=\linewidth]{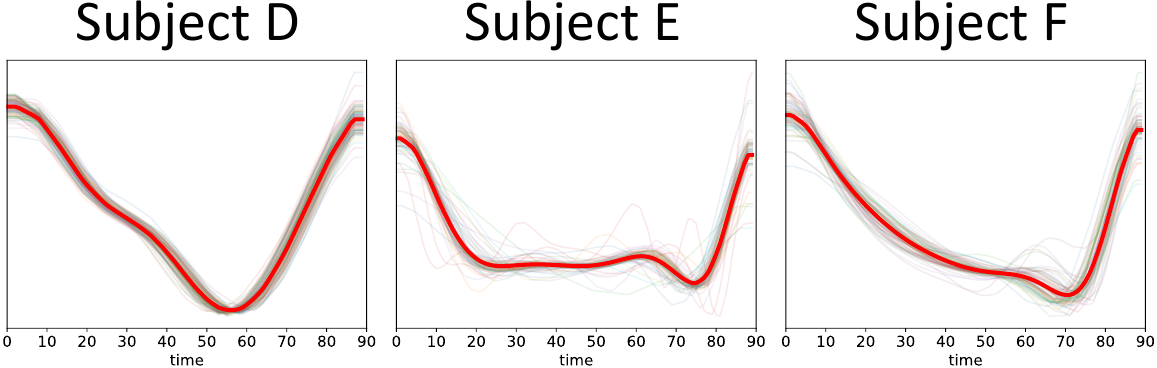}}
  \centerline{(b) rPPG periodic segments from UBFC-rPPG dataset}
\end{minipage}

\begin{minipage}[b]{0.95\linewidth}
  \centering
  \centerline{\includegraphics[width=\linewidth]{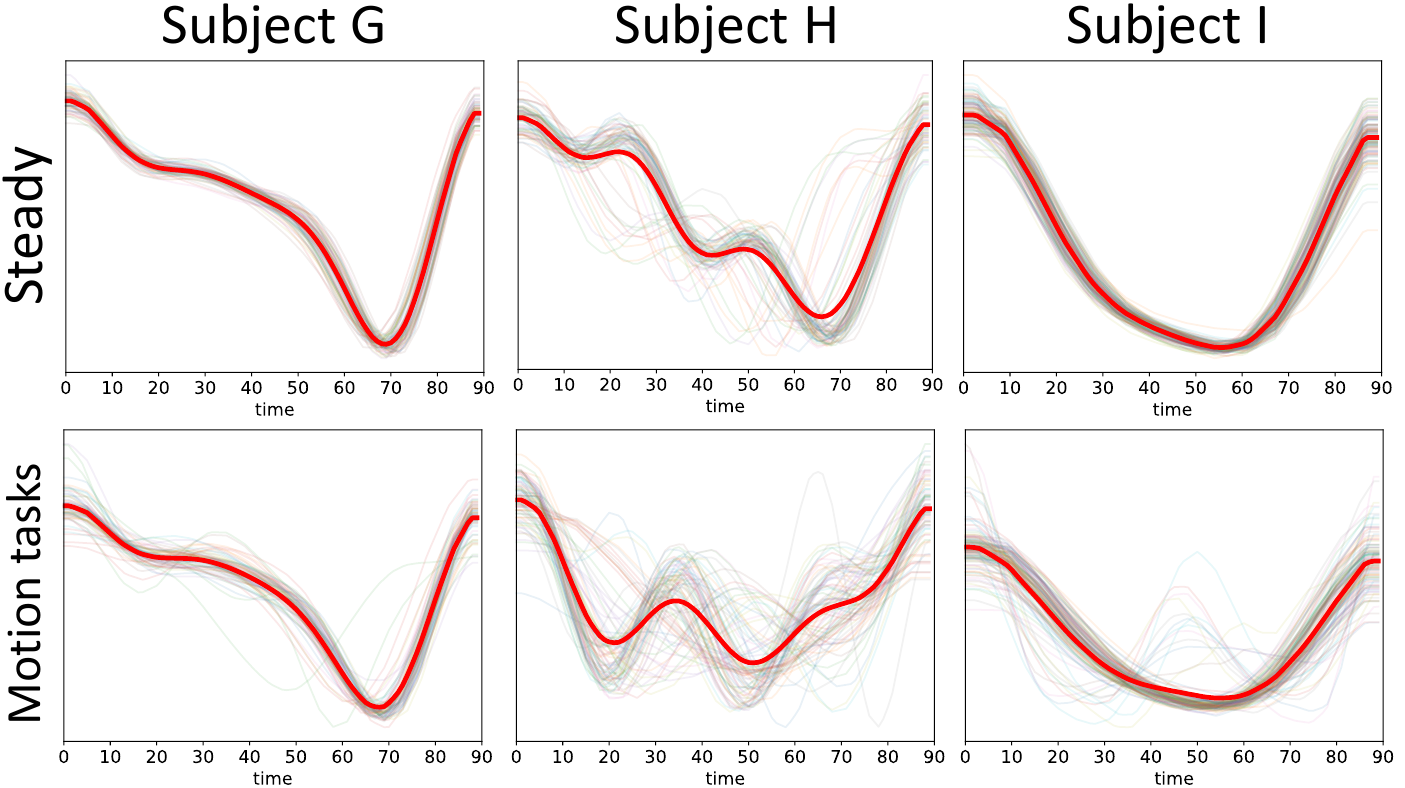}}
  \centerline{(c) rPPG periodic segments from PURE dataset}
\end{minipage}
\caption{rPPG periodic segments from (a) OBF dataset, (b) UBFC-rPPG dataset, and (c) PURE dataset. The red curves are the means of periodic segments.}
\label{fig:rppg_period_seg}
\vspace{-0.3cm}
\end{figure}

\begin{figure}[t]
    \centering
    \includegraphics[width=0.9\linewidth]{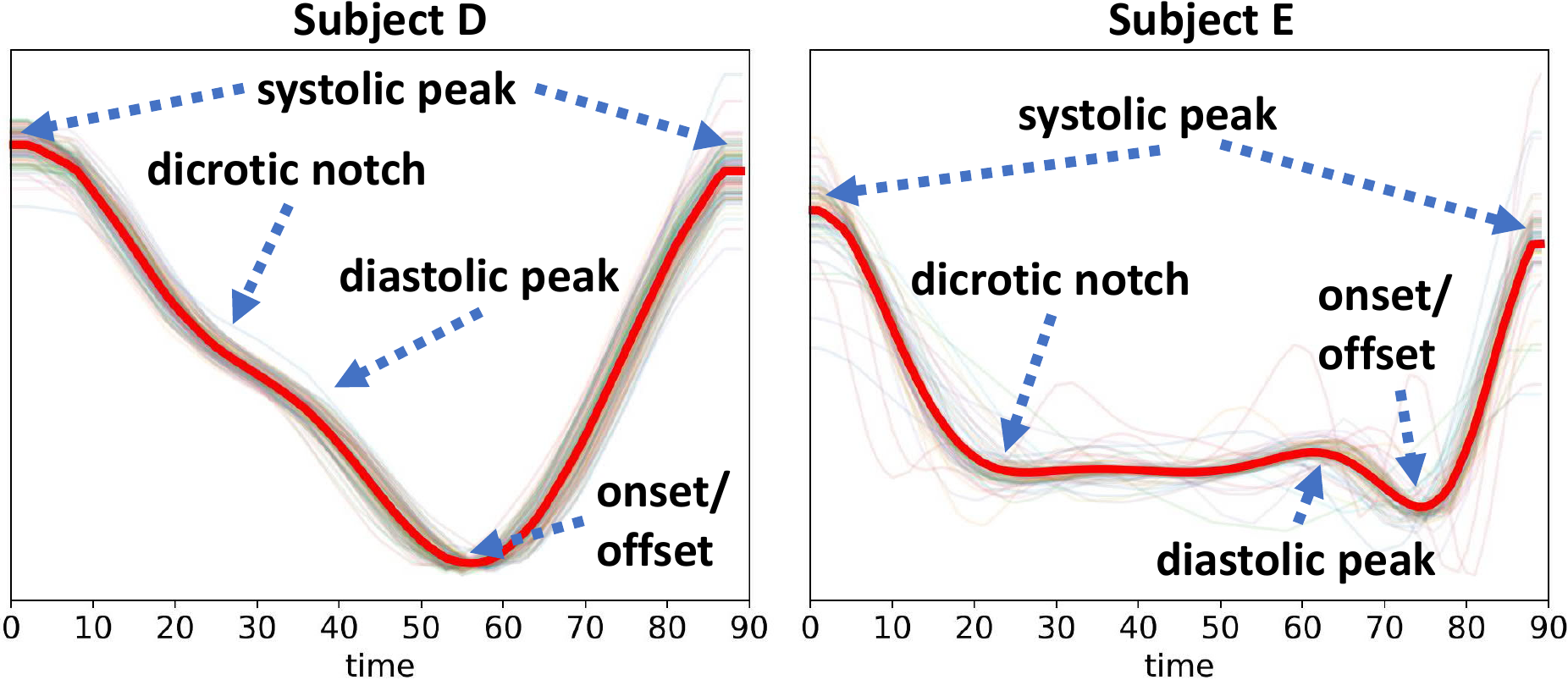}
    \caption{rPPG periodic segments and fiducial points from two subjects.}
    \label{fig:fiducial}
    \vspace{-0.3cm}
\end{figure}

The observed rPPG periodic segments from different subjects (subject A-I) in Fig. \ref{fig:rppg_period_seg} align with the aforementioned quantitative results. The rPPG periodic segments from the OBF dataset exhibit consistent morphology before and after exercises in Fig. \ref{fig:rppg_period_seg}(a). Conversely, the motion tasks in the PURE dataset significantly alter morphology in Fig. \ref{fig:rppg_period_seg}(c), resulting in noisy rPPG signals and a drop in performance during cross-session testing. Furthermore, the rPPG periodic segments from all three datasets display distinct morphologies for different subjects, highlighting the discriminative power of rPPG morphology. Fig. \ref{fig:fiducial} shows the subject-specific biometric characteristics of rPPG morphology in detail. The rPPG periodic segments from two subjects have distinct fiducial points \cite{lovisotto2020seeing} such as the systolic peaks, diastolic peaks, dicrotic notch, and onset/offset, which contain identity information.

Regarding fairness, prior studies \cite{nowara2020meta,vilesov2022blending} highlighted skin bias in rPPG signal quality. Dark skin may yield lower-quality rPPG signals, impacting authentication performance. We assess authentication performance for light and dark skin groups in the OBF dataset with a 20-heartbeat signal length and cross-session testing. For light skin, EER and AUC are 2.52\% and 97.79\%, respectively. For dark skin, EER and AUC are 4.04\% and 96.74\%. The performance of dark skin slightly falls behind that of light skin, indicating a skin tone bias in rPPG biometrics. Addressing this fairness issue may involve collecting more data from dark-skinned people or developing new algorithms, which remains a topic for future research.

\begin{table*}[htb!]
\resizebox{\linewidth}{!}
{
\begin{threeparttable}
\begin{tabular}{lccccc}
\toprule
\multirow{3}{*}{Biometric Methods}  & \multicolumn{5}{c}{EER$\downarrow$/AUC$\uparrow$} \\ \cmidrule(lr){2-6}  
                                    & \multicolumn{2}{c}{OBF}                        & UBFC-rPPG            & \multicolumn{2}{c}{PURE}                     \\ \cmidrule(lr){2-3} \cmidrule(lr){4-4} \cmidrule(lr){5-6} 
                                    & intra-sess           & cross-sess              & intra-sess           & intra-sess         & cross-sess              \\  \cmidrule{1-1} \cmidrule(lr){2-3} \cmidrule(lr){4-4} \cmidrule(lr){5-6}
FaceNet \cite{schroff2015facenet}$\blacktriangle$                             & 32.07\%/65.87\%      & 36.58\%/60.84\%         & 36.15\%/61.03\%      & 31.67\%/66.67\%    & 35.67\%/65.11\%         \\
Privacy-preserving FR \cite{ji2022privacy}$\blacktriangle$ & 6.46\%/91.24\%      & 6.52\%/91.92\%         & 7.26\%/90.25\%      & 6.88\%/91.27\%    & 7.82\%/90.77\% \\
Hwang2021 \cite{hwang2021variation}$\blacklozenge$    & 1.21\%/99.30\%       & 16.72\%/84.74\%         & \underline{6.30\%/94.02\%} & \bf 0\%/100\% & \bf 4.23\%/98.14\% \\
Patil2018 \cite{patil2018non}$\bigstar$                               & 14.97\%/89.42\%      & 39.79\%/62.14\%         & 8.53\%/88.70\%       & \underline{4.00\%/92.00\%} & 32.68\%/72.11\%         \\
Ours w/ rPPG training$\bigstar$             & \bf 0\%/100\%   & \underline{3.23\%/96.92\%}    & $-*$                    & \bf 0\%/100\% & 11.68\%/92.61\%         \\
\bf Ours w/ rPPG-cPPG hybrid training$\bigstar$ & \underline{0.17\%/99.97\%} &  \bf{2.16\%/98.10\%} & \bf 0\%/100\%   & \bf 0\%/100\% & \underline{9.59\%/93.70\%} \\
 \bottomrule
\end{tabular}
\begin{tablenotes}
\item $\blacktriangle$: face recognition (FR), $\blacklozenge$: cPPG biometrics, $\bigstar$: rPPG biometrics, $*$: Training does not converge.
\end{tablenotes}
\end{threeparttable}
}
\caption{Performance comparison between biometric methods including face recognition, cPPG biometrics, and rPPG biometrics. Note that de-identified videos proposed in the paper are used for face recognition and rPPG biometrics.}
\vspace{-0.3cm}
\label{tab:comparison}
\end{table*}

\subsubsection{Comparison with other biometrics.}

In Table \ref{tab:comparison}, we compare rPPG biometrics with related biometric methods, including face and cPPG biometrics, when the signal length is 20 beats. For face recognition, we choose the highly cited face recognition method (FaceNet \cite{schroff2015facenet}) to prove how general face recognition works on de-identified facial videos. We use FaceNet to extract embeddings from de-identified images and train two fully connected layers on the embeddings to get the classification results. Table \ref{tab:comparison} demonstrates that FaceNet \cite{schroff2015facenet} fails to work on de-identified videos, indicating that there is no facial appearance information in the de-identified videos. Since our rPPG biometric method is privacy-preserving for facial appearances, we also compare our method with the recent privacy-preserving face recognition \cite{ji2022privacy}. The results show that our method can achieve better performance than privacy-preserving face recognition \cite{ji2022privacy}. Our rPPG biometric authentication completely gets rid of facial appearance while the privacy-preserving face recognition \cite{ji2022privacy} only adds noises to partially remove facial appearances to guarantee face recognition performance, which may still have risks of privacy leakage. In addition, we also compare our method w/ rPPG-cPPG hybrid training to our method w/ rPPG training (only rPPG branch is used for training in Fig. \ref{fig:hybrid}, and the cPPG branch is disabled during training).



On the OBF dataset, ours w/ rPPG-cPPG hybrid training achieves similar intra-session performance to ours w/ rPPG training, but achieves the best cross-session performance. This means \emph{external} cPPG datasets introducing morphology information can improve generalization, such as cross-session performance. Furthermore, our rPPG biometrics exhibits better performance than cPPG biometrics \cite{hwang2021variation}. This is primarily because rPPG signals are extracted from both spatial and temporal representations, allowing for the utilization of more information compared to cPPG signals, which are measured from a single spatial point in the temporal dimension. However, this holds true only when the rPPG signals are of high quality.

On the UBFC-rPPG dataset, ours w/ rPPG-cPPG hybrid training achieves 100\% AUC but ours w/ rPPG training does not converge. The reason might be that it is difficult for the model to learn rPPG morphology from the small-scale UBFC-rPPG dataset without the help of the \textit{external} cPPG dataset. This suggests that the \textit{external} cPPG dataset can help the model to learn discriminative rPPG morphology information. Moreover, the performance of cPPG biometrics is still lower than that of our rPPG biometrics.

On the PURE dataset, both rPPG and cPPG biometrics demonstrate good performance in intra-session testing. However, in cross-session testing, our rPPG biometrics are surpassed by cPPG biometrics. This is likely due to significant facial motions in the test videos, which negatively impact the quality of rPPG signals and morphology, as shown in Figure \ref{fig:rppg_period_seg}(c). On the other hand, cPPG signals measured from fingertips are less affected by facial motions, allowing for better performance in this scenario.

\subsubsection{Results and discussions about rPPG morphology.}

 We also made an interesting finding that the rPPG-cPPG hybrid training can significantly improve rPPG morphology reconstruction. Table \ref{tab:ps} shows the Pearson correlations between the mean periodic segments of GT cPPG and rPPG. High Pearson correlations mean rPPG morphology better resembles the corresponding GT cPPG. Note that our method does not require any GT cPPG for rPPG morphology reconstruction, so we choose unsupervised rPPG methods including POS \cite{wang2016algorithmic}, ICA \cite{poh2010advancements}, and \cite{gideon2021way} for comparison. Ours w/ rPPG-cPPG hybrid training achieves significantly higher Pearson correlation than the baseline methods, CP2D, and ours w/ rPPG training, as the \emph{external} cPPG datasets introduce helpful morphology information via the hybrid training to refine the rPPG morphology. Such cPPG datasets are publicly available, and thus do not introduce extra costs of data collection.

\begin{table}[t]
\resizebox{\linewidth}{!}
{
\begin{tabular}{lc}
\hline
Methods                                     & Pearson Correlations$\uparrow$          \\ \hline
POS \cite{wang2016algorithmic}$*$                                       & 0.78                         \\
ICA \cite{poh2010advancements}$*$                                       & 0.77                         \\
Gideon2021 \cite{gideon2021way}$*$                                & 0.77                         \\ \hline
\rowcolor[HTML]{EFEFEF} 
\multicolumn{2}{c}{\cellcolor[HTML]{EFEFEF} \textit{After the 1st training stage}}         \\
\rowcolor[HTML]{EFEFEF} 
CP2D                                        & 0.78                         \\ \hline
\rowcolor[HTML]{C0C0C0} 
\multicolumn{2}{c}{\cellcolor[HTML]{C0C0C0} \textit{After the 1st and 2nd training stages}} \\
\rowcolor[HTML]{C0C0C0} 
Ours w/ rPPG training                       & 0.70                         \\
\rowcolor[HTML]{C0C0C0} 
\bf Ours w/ rPPG-cPPG hybrid training       & \textbf{0.87}                \\ \hline
\end{tabular}
}
\caption{Pearson correlations between GT cPPG periodic segments and the rPPG periodic segments.}
\vspace{-0.3cm}
\label{tab:ps}
\end{table}


\section{Conclusion}
In this paper, we validated the feasibility of rPPG biometrics from facial videos. We proposed a two-stage training scheme and novel cPPG-rPPG hybrid training by using \emph{external} cPPG biometric datasets to improve rPPG biometric authentication. Our method achieves good performance on both rPPG biometrics authentication and rPPG morphology reconstruction. In addition, our method uses de-identified facial videos for authentication, which can protect sensitive facial appearance information. Future work will focus on collecting a large-scale rPPG biometric dataset and studying influencing factors like temporal stability, lighting, and recording devices.

\section*{Acknowledgments}
This work was supported by the Research Council of Finland (former Academy of Finland) Academy Professor project EmotionAI (grants 336116, 345122), ICT 2023 project TrustFace (grant 345948), the University of Oulu \& Research Council of Finland Profi 7 (grant 352788), and by Infotech Oulu. The work was also supported by the Spearhead project 'Gaze on Lips' funded by the Eudaimonia Institute of the University of Oulu, Finland. The authors also acknowledge CSC-IT Center for Science, Finland, for providing computational resources.

{\small
\bibliographystyle{ieee}
\bibliography{egbib}

\begin{thebibliography}{10}\itemsep=-1pt

\bibitem{baltrusaitis2018openface}
T.~Baltrusaitis, A.~Zadeh, Y.~C. Lim, and L.-P. Morency.
\newblock Openface 2.0: Facial behavior analysis toolkit.
\newblock In {\em 2018 13th IEEE international conference on automatic face \& gesture recognition (FG 2018)}, pages 59--66. IEEE, 2018.

\bibitem{biswas2019cornet}
D.~Biswas, L.~Everson, M.~Liu, M.~Panwar, B.-E. Verhoef, S.~Patki, C.~H. Kim, A.~Acharyya, C.~Van~Hoof, M.~Konijnenburg, et~al.
\newblock Cornet: Deep learning framework for ppg-based heart rate estimation and biometric identification in ambulant environment.
\newblock {\em IEEE transactions on biomedical circuits and systems}, 2019.

\bibitem{bobbia2019unsupervised}
S.~Bobbia, R.~Macwan, Y.~Benezeth, A.~Mansouri, and J.~Dubois.
\newblock Unsupervised skin tissue segmentation for remote photoplethysmography.
\newblock {\em Pattern Recognition Letters}, 124:82--90, 2019.

\bibitem{chen2018deepphys}
W.~Chen and D.~McDuff.
\newblock Deepphys: Video-based physiological measurement using convolutional attention networks.
\newblock In {\em ECCV}, pages 349--365, 2018.

\bibitem{daugman2009iris}
J.~Daugman.
\newblock How iris recognition works.
\newblock In {\em The essential guide to image processing}, pages 715--739. Elsevier, 2009.

\bibitem{de2013robust}
G.~De~Haan and V.~Jeanne.
\newblock Robust pulse rate from chrominance-based rppg.
\newblock {\em IEEE Transactions on Biomedical Engineering}, 60(10):2878--2886, 2013.

\bibitem{du2023dual}
J.~Du, S.-Q. Liu, B.~Zhang, and P.~C. Yuen.
\newblock Dual-bridging with adversarial noise generation for domain adaptive rppg estimation.
\newblock In {\em CVPR}, 2023.

\bibitem{gideon2021way}
J.~Gideon and S.~Stent.
\newblock The way to my heart is through contrastive learning: Remote photoplethysmography from unlabelled video.
\newblock In {\em ICCV}, pages 3995--4004, 2021.

\bibitem{goldberger2000physiobank}
A.~L. Goldberger, L.~A. Amaral, L.~Glass, J.~M. Hausdorff, P.~C. Ivanov, R.~G. Mark, J.~E. Mietus, G.~B. Moody, C.-K. Peng, and H.~E. Stanley.
\newblock Physiobank, physiotoolkit, and physionet: components of a new research resource for complex physiologic signals.
\newblock {\em circulation}, 2000.

\bibitem{gu2003novel}
Y.~Gu, Y.~Zhang, and Y.~Zhang.
\newblock A novel biometric approach in human verification by photoplethysmographic signals.
\newblock In {\em 4th International IEEE EMBS Special Topic Conference on Information Technology Applications in Biomedicine, 2003.} IEEE, 2003.

\bibitem{hwang2021variation}
D.~Y. Hwang, B.~Taha, and D.~Hatzinakos.
\newblock Variation-stable fusion for ppg-based biometric system.
\newblock In {\em ICASSP}. IEEE, 2021.

\bibitem{hwang2020evaluation}
D.~Y. Hwang, B.~Taha, D.~S. Lee, and D.~Hatzinakos.
\newblock Evaluation of the time stability and uniqueness in ppg-based biometric system.
\newblock {\em IEEE Transactions on Information Forensics and Security}, 2020.

\bibitem{ji2022privacy}
J.~Ji, H.~Wang, Y.~Huang, J.~Wu, X.~Xu, S.~Ding, S.~Zhang, L.~Cao, and R.~Ji.
\newblock Privacy-preserving face recognition with learnable privacy budgets in frequency domain.
\newblock In {\em European Conference on Computer Vision}, pages 475--491. Springer, 2022.

\bibitem{karlen2013multiparameter}
W.~Karlen, S.~Raman, J.~M. Ansermino, and G.~A. Dumont.
\newblock Multiparameter respiratory rate estimation from the photoplethysmogram.
\newblock {\em IEEE Transactions on Biomedical Engineering}, 2013.

\bibitem{lee2020cross}
E.~Lee, A.~Ho, Y.-T. Wang, C.-H. Huang, and C.-Y. Lee.
\newblock Cross-domain adaptation for biometric identification using photoplethysmogram.
\newblock In {\em ICASSP}. IEEE, 2020.

\bibitem{li2023learning}
J.~Li, Z.~Yu, and J.~Shi.
\newblock Learning motion-robust remote photoplethysmography through arbitrary resolution videos.
\newblock In {\em AAAI}, 2023.

\bibitem{li2018obf}
X.~Li, I.~Alikhani, J.~Shi, T.~Seppanen, J.~Junttila, K.~Majamaa-Voltti, M.~Tulppo, and G.~Zhao.
\newblock The obf database: A large face video database for remote physiological signal measurement and atrial fibrillation detection.
\newblock In {\em 2018 13th IEEE International Conference on Automatic Face \& Gesture Recognition (FG 2018)}, pages 242--249. IEEE, 2018.

\bibitem{li2014remote}
X.~Li, J.~Chen, G.~Zhao, and M.~Pietikainen.
\newblock Remote heart rate measurement from face videos under realistic situations.
\newblock In {\em CVPR}, pages 4264--4271, 2014.

\bibitem{liu2022learning}
S.-Q. Liu, X.~Lan, and P.~C. Yuen.
\newblock Learning temporal similarity of remote photoplethysmography for fast 3d mask face presentation attack detection.
\newblock {\em IEEE Transactions on Information Forensics and Security}, 2022.

\bibitem{NEURIPS2020_e1228be4}
X.~Liu, J.~Fromm, S.~Patel, and D.~McDuff.
\newblock Multi-task temporal shift attention networks for on-device contactless vitals measurement.
\newblock In H.~Larochelle, M.~Ranzato, R.~Hadsell, M.~F. Balcan, and H.~Lin, editors, {\em NeurIPS}, volume~33, pages 19400--19411, 2020.

\bibitem{liu2018learning}
Y.~Liu, A.~Jourabloo, and X.~Liu.
\newblock Learning deep models for face anti-spoofing: Binary or auxiliary supervision.
\newblock In {\em CVPR}, 2018.

\bibitem{lovisotto2020seeing}
G.~Lovisotto, H.~Turner, S.~Eberz, and I.~Martinovic.
\newblock Seeing red: Ppg biometrics using smartphone cameras.
\newblock In {\em CVPRW}, 2020.

\bibitem{lu2021dual}
H.~Lu, H.~Han, and S.~K. Zhou.
\newblock Dual-gan: Joint bvp and noise modeling for remote physiological measurement.
\newblock In {\em CVPR}, pages 12404--12413, 2021.

\bibitem{lu2023neuron}
H.~Lu, Z.~Yu, X.~Niu, and Y.-C. Chen.
\newblock Neuron structure modeling for generalizable remote physiological measurement.
\newblock In {\em CVPR}, pages 18589--18599, 2023.

\bibitem{luque2018end}
J.~Luque, G.~Cortes, C.~Segura, A.~Maravilla, J.~Esteban, and J.~Fabregat.
\newblock End-to-end photopleth ysmography (ppg) based biometric authentication by using convolutional neural networks.
\newblock In {\em 2018 26th European Signal Processing Conference (EUSIPCO)}. IEEE, 2018.

\bibitem{mcduff2014morph}
D.~McDuff, S.~Gontarek, and R.~W. Picard.
\newblock Remote detection of photoplethysmographic systolic and diastolic peaks using a digital camera.
\newblock {\em IEEE Transactions on Biomedical Engineering}, 2014.

\bibitem{niu2018synrhythm}
X.~Niu, H.~Han, S.~Shan, and X.~Chen.
\newblock Synrhythm: Learning a deep heart rate estimator from general to specific.
\newblock In {\em ICPR}, pages 3580--3585. IEEE, 2018.

\bibitem{niu2019rhythmnet}
X.~Niu, S.~Shan, H.~Han, and X.~Chen.
\newblock Rhythmnet: End-to-end heart rate estimation from face via spatial-temporal representation.
\newblock {\em IEEE Transactions on Image Processing}, 29:2409--2423, 2019.

\bibitem{niu2020video}
X.~Niu, Z.~Yu, H.~Han, X.~Li, S.~Shan, and G.~Zhao.
\newblock Video-based remote physiological measurement via cross-verified feature disentangling.
\newblock In {\em ECCV}, pages 295--310. Springer, 2020.

\bibitem{nowara2020meta}
E.~M. Nowara, D.~McDuff, and A.~Veeraraghavan.
\newblock A meta-analysis of the impact of skin tone and gender on non-contact photoplethysmography measurements.
\newblock In {\em Proceedings of the IEEE/CVF Conference on Computer Vision and Pattern Recognition Workshops}, 2020.

\bibitem{nowara2021benefit}
E.~M. Nowara, D.~McDuff, and A.~Veeraraghavan.
\newblock The benefit of distraction: Denoising camera-based physiological measurements using inverse attention.
\newblock In {\em ICCV}, pages 4955--4964, 2021.

\bibitem{patil2018non}
O.~R. Patil, W.~Wang, Y.~Gao, W.~Xu, and Z.~Jin.
\newblock A non-contact ppg biometric system based on deep neural network.
\newblock In {\em 2018 IEEE 9th International Conference on Biometrics Theory, Applications and Systems (BTAS)}. IEEE, 2018.

\bibitem{pimentel2016toward}
M.~A. Pimentel, A.~E. Johnson, P.~H. Charlton, D.~Birrenkott, P.~J. Watkinson, L.~Tarassenko, and D.~A. Clifton.
\newblock Toward a robust estimation of respiratory rate from pulse oximeters.
\newblock {\em IEEE Transactions on Biomedical Engineering}, 2016.

\bibitem{poh2010advancements}
M.-Z. Poh, D.~J. McDuff, and R.~W. Picard.
\newblock Advancements in noncontact, multiparameter physiological measurements using a webcam.
\newblock {\em IEEE transactions on biomedical engineering}, 58(1):7--11, 2010.

\bibitem{schroff2015facenet}
F.~Schroff, D.~Kalenichenko, and J.~Philbin.
\newblock Facenet: A unified embedding for face recognition and clustering.
\newblock In {\em CVPR}, pages 815--823, 2015.

\bibitem{speth2023non}
J.~Speth, N.~Vance, P.~Flynn, and A.~Czajka.
\newblock Non-contrastive unsupervised learning of physiological signals from video.
\newblock In {\em CVPR}, 2023.

\bibitem{vspetlik2018visual}
R.~{\v{S}}petl{\'\i}k, V.~Franc, and J.~Matas.
\newblock Visual heart rate estimation with convolutional neural network.
\newblock In {\em BMVC}, pages 3--6, 2018.

\bibitem{stricker2014non}
R.~Stricker, S.~M{\"u}ller, and H.-M. Gross.
\newblock Non-contact video-based pulse rate measurement on a mobile service robot.
\newblock In {\em The 23rd IEEE International Symposium on Robot and Human Interactive Communication}, pages 1056--1062. IEEE, 2014.

\bibitem{sun2022contrast}
Z.~Sun and X.~Li.
\newblock Contrast-phys: Unsupervised video-based remote physiological measurement via spatiotemporal contrast.
\newblock In {\em ECCV}, pages 492--510. Springer, 2022.

\bibitem{tulyakov2016self}
S.~Tulyakov, X.~Alameda-Pineda, E.~Ricci, L.~Yin, J.~F. Cohn, and N.~Sebe.
\newblock Self-adaptive matrix completion for heart rate estimation from face videos under realistic conditions.
\newblock In {\em CVPR}, pages 2396--2404, 2016.

\bibitem{verkruysse2008remote}
W.~Verkruysse, L.~O. Svaasand, and J.~S. Nelson.
\newblock Remote plethysmographic imaging using ambient light.
\newblock {\em Optics express}, 16(26):21434--21445, 2008.

\bibitem{vilesov2022blending}
A.~Vilesov, P.~Chari, A.~Armouti, A.~B. Harish, K.~Kulkarni, A.~Deoghare, L.~Jalilian, and A.~Kadambi.
\newblock Blending camera and 77 ghz radar sensing for equitable, robust plethysmography.
\newblock {\em ACM Trans. Graph.(SIGGRAPH)}, 2022.

\bibitem{wang2022self}
H.~Wang, E.~Ahn, and J.~Kim.
\newblock Self-supervised representation learning framework for remote physiological measurement using spatiotemporal augmentation loss.
\newblock {\em AAAI}, 2022.

\bibitem{wang2016algorithmic}
W.~Wang, A.~C. den Brinker, S.~Stuijk, and G.~De~Haan.
\newblock Algorithmic principles of remote ppg.
\newblock {\em IEEE Transactions on Biomedical Engineering}, 64(7):1479--1491, 2016.

\bibitem{wang2014exploiting}
W.~Wang, S.~Stuijk, and G.~De~Haan.
\newblock Exploiting spatial redundancy of image sensor for motion robust rppg.
\newblock {\em IEEE transactions on Biomedical Engineering}, 62(2):415--425, 2014.

\bibitem{wildes1997iris}
R.~P. Wildes.
\newblock Iris recognition: an emerging biometric technology.
\newblock {\em Proceedings of the IEEE}, 85(9):1348--1363, 1997.

\bibitem{yang2022simper}
Y.~Yang, X.~Liu, J.~Wu, S.~Borac, D.~Katabi, M.-Z. Poh, and D.~McDuff.
\newblock Simper: Simple self-supervised learning of periodic targets.
\newblock In {\em ICLR}, 2022.

\bibitem{yao2007pilot}
J.~Yao, X.~Sun, and Y.~Wan.
\newblock A pilot study on using derivatives of photoplethysmographic signals as a biometric identifier.
\newblock In {\em 2007 29th Annual International Conference of the IEEE Engineering in Medicine and Biology Society}. IEEE, 2007.

\bibitem{yu2019remoteBMVC}
Z.~Yu, X.~Li, and G.~Zhao.
\newblock Remote photoplethysmograph signal measurement from facial videos using spatio-temporal networks.
\newblock In {\em BMVC}, page 277. {BMVA} Press, 2019.

\bibitem{yu2019remote}
Z.~Yu, W.~Peng, X.~Li, X.~Hong, and G.~Zhao.
\newblock Remote heart rate measurement from highly compressed facial videos: an end-to-end deep learning solution with video enhancement.
\newblock In {\em ICCV}, pages 151--160, 2019.

\bibitem{yu2023physformer++}
Z.~Yu, Y.~Shen, J.~Shi, H.~Zhao, Y.~Cui, J.~Zhang, P.~Torr, and G.~Zhao.
\newblock Physformer++: Facial video-based physiological measurement with slowfast temporal difference transformer.
\newblock {\em International Journal of Computer Vision}, 2023.

\bibitem{yu2022physformer}
Z.~Yu, Y.~Shen, J.~Shi, H.~Zhao, P.~H. Torr, and G.~Zhao.
\newblock Physformer: facial video-based physiological measurement with temporal difference transformer.
\newblock In {\em CVPR}, pages 4186--4196, 2022.

\bibitem{yue2023facial}
Z.~Yue, M.~Shi, and S.~Ding.
\newblock Facial video-based remote physiological measurement via self-supervised learning.
\newblock {\em TPAMI}, 2023.

\end{thebibliography}
}

\end{document}